\pdfoutput=1
\documentclass[11pt]{article}

\usepackage[]{ACL2023}

\usepackage{times}
\usepackage{latexsym}

\usepackage[T1]{fontenc}
\usepackage[utf8]{inputenc}

\usepackage{microtype}

\usepackage{inconsolata}
\usepackage{array}

\usepackage{natbib}

\usepackage{amsmath,amsfonts}
\usepackage{graphicx}
\usepackage{subcaption}
\usepackage[normalem]{ulem}

\usepackage{graphicx}
\usepackage{subcaption}
\usepackage{tikz}

\usepackage{algorithm}
\usepackage{algpseudocode}

\usepackage{xcolor}
\usepackage{booktabs}
\usepackage{multirow}
\usepackage{multicol}

\usepackage{adjustbox}
\usepackage{array} % Required for new column type

\newcolumntype{C}[1]{>{\centering\arraybackslash}m{#1}}

\makeatletter
\def\@fnsymbol#1{}
\makeatother

\title{
Deep Learning for Network Anomaly Detection under Data Contamination: Evaluating Robustness and Mitigating Performance Degradation
}

\author{\fontsize{10}{10}\selectfont D'Jeff K. Nkashama$^{1,2}$ Jordan Masakuna Félicien$^{1,2}$ Arian Soltani$^{1,2}$ Jean-Charles Verdier$^{1,2}$ \\{\fontsize{10}{10}\selectfont \bf Pierre-Martin Tardif$^{1,3}$ Marc Frappier$^{1,2}$ Froduald Kabanza$^{1,2}$}  
\thanks{$^1$Groupe de Recherche Interdisciplinaire en Cybersécurité (GRIC), $^{2}$Department of Computer Science, Université de Sherbrooke, Canada $^{3}$Department of Management School, Université de Sherbrooke, Canada. Correspondence email:\href{mailto:djeff.nkashama.kanda@usherbrooke.ca}{\texttt{djeff.nkashama.kanda@usherbrooke.ca}}}
} 

\date{}

\begin{document}

\maketitle

\begin{abstract}
%Deep learning (DL) has become a key tool for securing computer networks, particularly in network anomaly detection (NAD). DL approaches have gained popularity in cybersecurity due to their ability to extract meaningful features and learn patterns and implicit rules directly from data. However, in NAD, DL models can be compromised if they are trained on data mistakenly assumed to be benign but that actually contains hidden attack-related data, known as data contamination. This paper evaluates the robustness of six unsupervised DL algorithms against data contamination using our proposed evaluation protocol. The experiments reveal that data contamination significantly degrades the performance of state-of-the-art anomaly detection algorithms. These results underscore the critical need for self-protection mechanisms against data contamination when developing novel deep learning-based NAD models. To mitigate this issue, we propose enhancing a classical auto-encoder, a commonly used DL model for network anomaly detection, by adding a constraint to its latent representation. This modification encourages normal data to concentrate more densely around a learnable center $c$ in the latent space. Our evaluation demonstrates that our proposed approach improves resistance to data contamination compared to existing methods.

Deep learning (DL) has emerged as a crucial tool in network anomaly detection (NAD) for cybersecurity. While DL models 
 for anomaly detection excel at extracting features and learning patterns from data, they are vulnerable to data contamination---the inadvertent inclusion of attack-related data in training sets presumed benign. This study evaluates the robustness of six unsupervised DL algorithms against data contamination using our proposed evaluation protocol. Results demonstrate significant performance degradation in state-of-the-art anomaly detection algorithms when exposed to contaminated data, highlighting the critical need for self-protection mechanisms in DL-based NAD models. To mitigate this vulnerability, we propose an enhanced auto-encoder with a constrained latent representation, allowing normal data to cluster more densely around a learnable center in the latent space. Our evaluation reveals that this approach exhibits improved resistance to data contamination compared to existing methods, offering a promising direction for more robust NAD systems.

\end{abstract}

\section{Introduction}
In today's digital age, information technology has become an integral part of our daily activities, including communication, travel, work, banking, education, and manufacturing. An illustrative example of this integration is the Internet of Things (IoT), marking a substantial change in our interaction with technology \cite{sarker2023internet}.  IoT allows remote monitoring and control, bridging the gap between the physical world and computer-based systems. While technology offers numerous benefits, it can also be exploited by adversaries to conduct malicious activities on cyberinfrastructure \cite{ferencz2021review,bovenzi2023network}. These malicious activities can take various forms, including information theft, denial of service, and data contamination, among others, posing threats to operational cyber systems. Consequently, the need to safeguard cyberinfrastructures against these persistent and destructive cyber threats is of paramount importance. 

Recent advances in machine learning (ML) or deep learning (DL) have played a fundamental role and made remarkable progress in the field of cybersecurity at a rapid pace \cite{kwon2019survey,liu2020anomaly,chen2018autoencoder,crawford2015survey}. Numerous ML-based anomaly detection (AD) techniques have found practical use in the realm of cybersecurity \cite{kwon2019survey,sharafaldin2018toward,chen2018autoencoder,crawford2015survey}. An anomaly represents an observation significantly deviating from what is considered normal behavior \cite{alvarez2022revealing}. Based on the context, such observations can be deemed unusual, irregular, inconsistent, unexpected, faulty, fraudulent, or malicious \cite{ruff2021unifying,chalapathy2019deep}. Examples of AD applications in cybersecurity include malware detection, spam filtering, and intrusion detection systems  (IDS) \cite{crawford2015survey,liu2020anomaly,nelson2008exploiting,xiao2015feature,dua2016data}.

Training ML models for AD requires a large amount of data usually collected from various sources (e.g. sensors, actuators, and processes). However, as the volume of data grows, it becomes increasingly vulnerable to contamination \cite{ilyas2019adversarial,bovenzi2023network,9838942,31906193190637}.  There are several potential sources of data contamination. One source of contamination could be an ongoing attack campaign, which may go unnoticed during data collection, leading to the inclusion of malicious instances in the training data. Misconfigurations in equipment are another source of contamination. 
Moreover, adversaries, aware of the importance of data in ML-based IDS, could attempt to deliberately inject malicious samples into the training set. 
%\cite{hayase2021spectre}\textemdash a tactic known as data contamination attack.
The adversaries' goal is to hamper model performance and possibly create a backdoor for subsequent intrusions.  For example, in a network traffic analysis within an IoT environment, data contamination  can cause a DL model to erroneously predict traffic related to a Denial of Service (DoS) attack as legitimate user activity \cite{nelson2008exploiting,xiao2015feature}. This misclassification has the potential to disrupt the availability of services. Furthermore, an attacker may compromise a device by gaining remote control and injecting carefully crafted traffic from an external network. 
While evasion attacks during testing \cite{biggio2018wild,nelson2008exploiting} are also a concern, our research emphasizes the importance of mitigating data contamination during the model's training process to enhance robustness against data contamination.

In the current landscape of AD model training, it is typically assumed that the training data is clean and free of contamination. However, ensuring the cleanliness of data can be a resource-intensive process, especially when dealing with large-scale datasets \cite{hayase2021spectre}. Many AD models are not routinely subjected to data verification procedures. This becomes particularly critical in intrusion detection, where the resilience of AD models against data contamination is crucial. Although the robustness of AD models has already been studied, the most recent AD models are rarely exposed to and tested on contaminated data \cite{qiu2021neural,bovenzi2023network}. This lack of exposure to real-world data contamination can be a significant limitation when considering the practicability of these models.
Effectively addressing this challenge involves developing robust AD models.

This paper evaluates the impact of training data contamination on the performance of six unsupervised deep learning models for network AD. We chose these models because they are recent deep learning-based AD methods, selected for their superior performance compared to their predecessors \cite{qiu2021neural,chen2018autoencoder,zong2018deep,zenati2018adversarially}, and the diversity in their approaches to modeling the AD task. These approaches include self-supervised \cite{qiu2021neural}, generative adversarial \cite{zenati2018adversarially}, energy-based \cite{zhai2016deep}, density modeling \cite{zong2018deep}, and reconstruction-based \cite{li2021deep,zong2018deep,chen2018autoencoder} methods. It is important to note that the data contamination discussed in this study involves injecting malicious instances into the training set, which is expected to be cleaned for  the AD task, while keeping their features unchanged.

\textbf{Contributions}:

In this paper, our contribution is threefold. Firstly, we propose an evaluation protocol that integrates data contamination to emulate a real-world challenge in training DL for unsupervised AD task. The protocol enables fair and rigorous assessment of models' robustness and prevents data leakage during the evaluation process. We employ this protocol to evaluate the robustness of six state-of-the-art DL models for unsupervised AD: ALAD \cite{zenati2018adversarially}, Deep auto-encoder \cite{chen2018autoencoder}, DAGMM \cite{zong2018deep}, DSEBM \cite{zhai2016deep}, DUAD \cite{li2021deep}, and NeuTraLAD \cite{qiu2021neural}. Through a series of experiments, we systematically analyze the performance of the selected models under various levels of data contamination. We use the following cybersecurity benchmark datasets: CIC-CSE-IDS2018, Kitsune, CIC-IoT23, KDDCUP, and NSL-KDD\footnote{\href{https://www.unb.ca/cic/datasets/nsl.html}{https://www.unb.ca/cic/datasets/nsl.html}}.

Secondly, we enhance the performance of a standard deep autoencoder (DAE) by adding a constraint that data in its latent space should be more densely concentrate around a learnable center $c$. We refer to the improved models as DAE-LR and DUAD-LR. Our experiments demonstrate significant improvements in overall performance, particularly when dealing with data contamination, compared to standard versions of autoencoders, a mostly used model in network AD \cite{9838942,9872039,bovenzi2023network}.

Lastly, through an extensive experiments, we underscore the potential pitfalls of relying on outdated cybersecurity datasets for evaluating model performance. We demonstrate how model performance on such datasets can be misleading when assessing their effectiveness in current network environments with evolving attack patterns. We also emphasize the significance of robustness to data contamination as a critical criterion when selecting an AD model for cybersecurity applications. Our study provides valuable guidance to practitioners and researchers in choosing models that can withstand data contamination commonly encountered in real-world intrusion detection scenarios.

\section{Related Work and Background}
\label{relwork}
\subsection{Deep Unsupervised Anomaly Detection}
Unsupervised anomaly detection consists of separating normal data from anomalies or learning the underlying data distribution of normal data with no access to labels. 
This approach is challenging since the absence of explicit labels limits exploitable information. 

Unsupervised Anomaly detection (UAD) approaches can be broadly categorized into four types. (1) Probabilistic methods, such as DAGMM \cite{zong2018deep}, estimate the probability density function of the data and flag samples lying in the low-density region as anomalies. (2) Reconstruction-based methods aim to approximate an identity function of input data, assuming that normal data is compressible and can be reconstructed accurately. Based on the reconstruction error\textemdash the scoring function\textemdash the learned model flags data samples that cannot be reconstructed well from their compression as anomalous. Examples of these methods include MemAE \cite{gong2019memorizing} and DUAD \cite{li2021deep}. (3) Distance-based models, such as LOF \cite{breunig2000lof} and iForest \cite{liu2008isolation}, predict anomalies by measuring their distances from normal data, with unusual samples being those with significantly different distances. (4) One-class classification approaches, such as OC-SVM \cite{scholkopf1999support} and DeepSVDD \cite{ruff2018deep}, try to identify normal data points amongst all data, primarily by learning the underlying data distribution of only normal data. Each of these categories has its own strengths and weaknesses, making them suitable for different types of AD tasks.

Classical approaches to unsupervised AD work well with lower dimensions. However, as the feature space grows, they suffer from the curse of dimensionality \cite{li2021deep}. Deep learning architectures, for instance, autoencoders \cite{chen2018autoencoder} have shown significant potential to mitigate this issue \cite{simmross2011anomaly,zong2018deep}.

In this paper, we evaluate the robustness of various unsupervised deep learning methods against data contamination. We selected these methods based on their recent developments, superior performance compared to their predecessors, and the diversity of their approaches to modeling the AD task, including reconstruction-based, energy-based, generative-based, and self-supervised-based approaches \cite{zenati2018adversarially,chen2018autoencoder,zong2018deep,zhai2016deep,li2021deep,qiu2021neural}. No such study, with a wide range of approaches, on network AD has been conducted  as far as we know, yet resilience to data contamination should be an important requirement of deep learning-based network AD methods. 
\subsection{Data contamination}
Training a network AD model requires data containing only benign traffic. However, in reality, collected data may contain both benign and malicious instances, resulting in a contaminated dataset. Contaminated instances may lead to higher prediction errors during testing and deployment \cite{awan2021contra}. There are several causes of these corrupted instances, including misconfiguration, ongoing attacks during data collection, or adversaries attempting to manipulate the training process to create backdoors. %Adversaries may pursue different goals with these attacks (which can be generic or specific). In a \textit{generic attack}, the adversary aims to increase misclassification without specifically targeting a particular class. The goal is to disrupt the model's overall performance, making it less effective in identifying instances correctly across all classes. In contrast, a \textit{specific attack} is when the adversary targets a particular class. The intention is to cause misclassification specifically within these classes, potentially for malicious purposes related to those classes. These distinctions help characterize the adversary's objectives and guide the development of defenses against such attacks.
 %%SOLVED
%Research on data contamination has gradually surged, leading to a variety of techniques. For instance, earlier works on SVMs \cite{biggio2012poisoning}, feature selection \cite{xiao2015feature}, and PCA \cite{rubinstein2009antidote}, as well as initial efforts with deep learning that mostly focused on decreasing false classification \cite{munoz2017towards}. Later, this area was dominated by research based on generative models \cite{goodfellow2014generative}. With generative models, attacks could aim toward creating a backdoor, where samples with specific and easy-to-manipulate characteristics would bypass detection by the model \cite{wang2020certifying}. All this work mostly studied data contamination in the case of images classification \cite{biggio2012poisoning,biggio2018wild, fang2020local}, linear regression \cite{jagielski2018manipulating}, IoT environment \cite{bovenzi2023network}, but little attention has been paid to data contamination in the context of network intrusion detection systems with recent and best performing, and wide range of deep learning models for network anomaly detection and on how to mitigate models performance degradation. In this study, we evaluate the robustness models against data contamination, and propose a latent regulated approach to improve the performance of deep autoencoder.
Research on data contamination has gradually surged, leading to the development of various techniques. Earlier works focused on SVMs \cite{biggio2012poisoning}, feature selection \cite{xiao2015feature}, and PCA \cite{rubinstein2009antidote}, as well as initial deep learning efforts that mainly addressed reducing false classification \cite{munoz2017towards}. Subsequently, research in this area shifted toward generative models \cite{goodfellow2014generative}, where attacks could create a backdoor, allowing samples with specific and easily manipulated characteristics to evade detection by the model \cite{wang2020certifying}. Most of these studies primarily focused on data contamination in image classification \cite{biggio2012poisoning,biggio2018wild,fang2020local}, linear regression \cite{jagielski2018manipulating}, and IoT environments \cite{bovenzi2023network}. However, little attention has been paid to data contamination in the context of network intrusion detection systems, especially considering the recent advancements and high-performing unsupervised deep learning models for AD, as well as the mitigation of model performance degradation.
In this study, we evaluate the robustness of models against data contamination and propose a latent regulated approach to improve the performance of deep autoencoders.

\section{Method}
\subsection{Latent Regulated Deep Auto-Encoder}

In Figure \ref{fig:deepae}, we present the architecture of the Deep Autoencoder (DAE), which consists of two components: an encoder and a decoder. The encoder's role is to transform the input data into a low-dimensional representation, referred to as the latent representation. In contrast, the decoder's function is to reconstruct the original input data from this latent representation. Training this model is done by minimizing the reconstruction error in Equation (\ref{eq:recerr}). Furthermore, the same reconstruction error also serves as the anomaly scoring function.
\begin{figure}[ht]
  \tikzset{arrow/.style={-stealth, thick, draw=gray!80!black}}
\centering
\begin{tikzpicture}[thick,scale=0.7, every node/.style={scale=0.7}]
%     \draw[help lines](0,-5) grid (10,5);  
     
	\node[fill=green!20, minimum width=0.5cm, minimum height=3.5cm] (X) at (0,0) {$\mathbf x$};
	
	\draw[fill=blue!25] ([xshift=0.5cm]X.north east) -- ([xshift=2.5cm,yshift=0.5cm]X.east) -- ([xshift=2.5cm,yshift=-0.5cm]X.east) -- ([xshift=0.5cm]X.south east) -- cycle; 
	\node at (1.75,0) {\textsc{Encoder}};
	
	\node[fill=purple!20, minimum width=0.5cm, minimum height=1.0cm] (Z) at (3.5cm,0) {$\mathbf z$};
	
	\draw[fill=blue!20] ([xshift=0.5cm]Z.north east) -- ([xshift=2.5cm,yshift=1.25cm]Z.north east) -- ([xshift=2.5cm,yshift=-1.25cm]Z.south east) -- ([xshift=0.5cm]Z.south east) -- cycle;
	\node at (5.25,0) {\textsc{Decoder}};
	
	\node[fill=red!20, minimum width=0.5cm, minimum height=3.5cm] (Xp) at (7,0) {$\mathbf{\hat{x}}$};
	
	\draw[arrow] (X.east) -- ([xshift=0.5cm]X.east);
	\draw[arrow] ([xshift=-0.5cm]Z.west) -- (Z.west);
	\draw[arrow] (Z.east) -- ([xshift=0.5cm]Z.east);
	\draw[arrow] ([xshift=-0.5cm]Xp.west) -- (Xp.west);
     
\end{tikzpicture}
\caption{Deep Auto-Encoder architecture where, $x$, $z$, and $x'$ denote the input data, the latent representation, and the reconstructed input, respectively.}
\label{fig:deepae}
\end{figure}
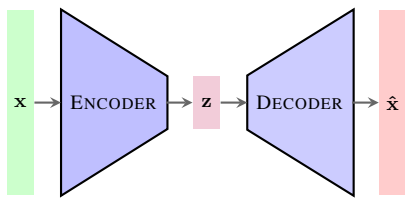
\begin{equation}
   \label{eq:recerr}
     L_\theta(x) = \|x - \hat{x}\|^2\,.
 \end{equation}
where, $\|.\|$ is the $\ell_2$-norm.
\begin{figure*}[h]
    \centering
    \begin{subfigure}[]{.3\textwidth}
        
        \includegraphics[height=3.5cm]{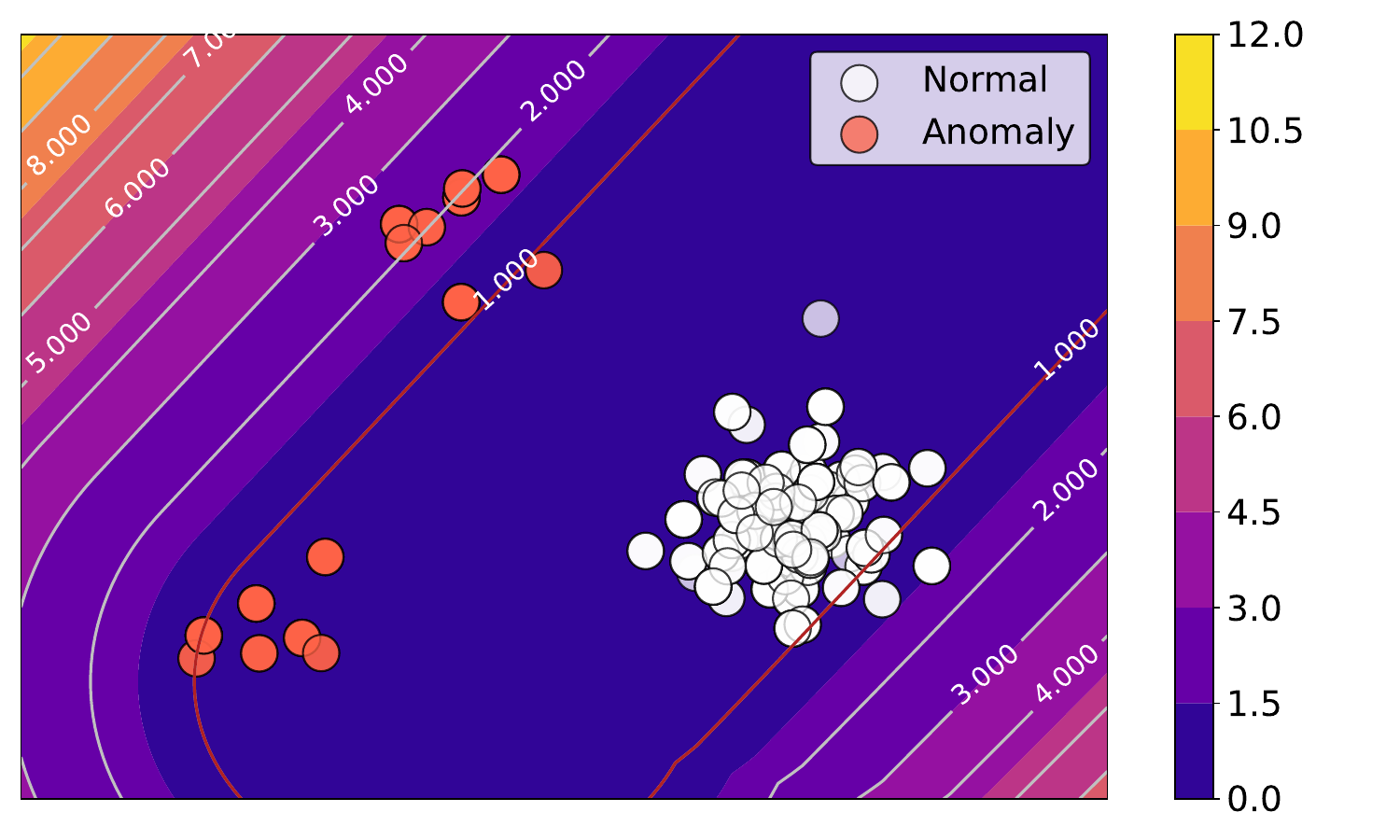}
        \caption{}
        \label{fig:toy_dae_standard}
    \end{subfigure}
    \hfill
    \begin{subfigure}[]{.3\textwidth}
        \includegraphics[height=3.5cm]{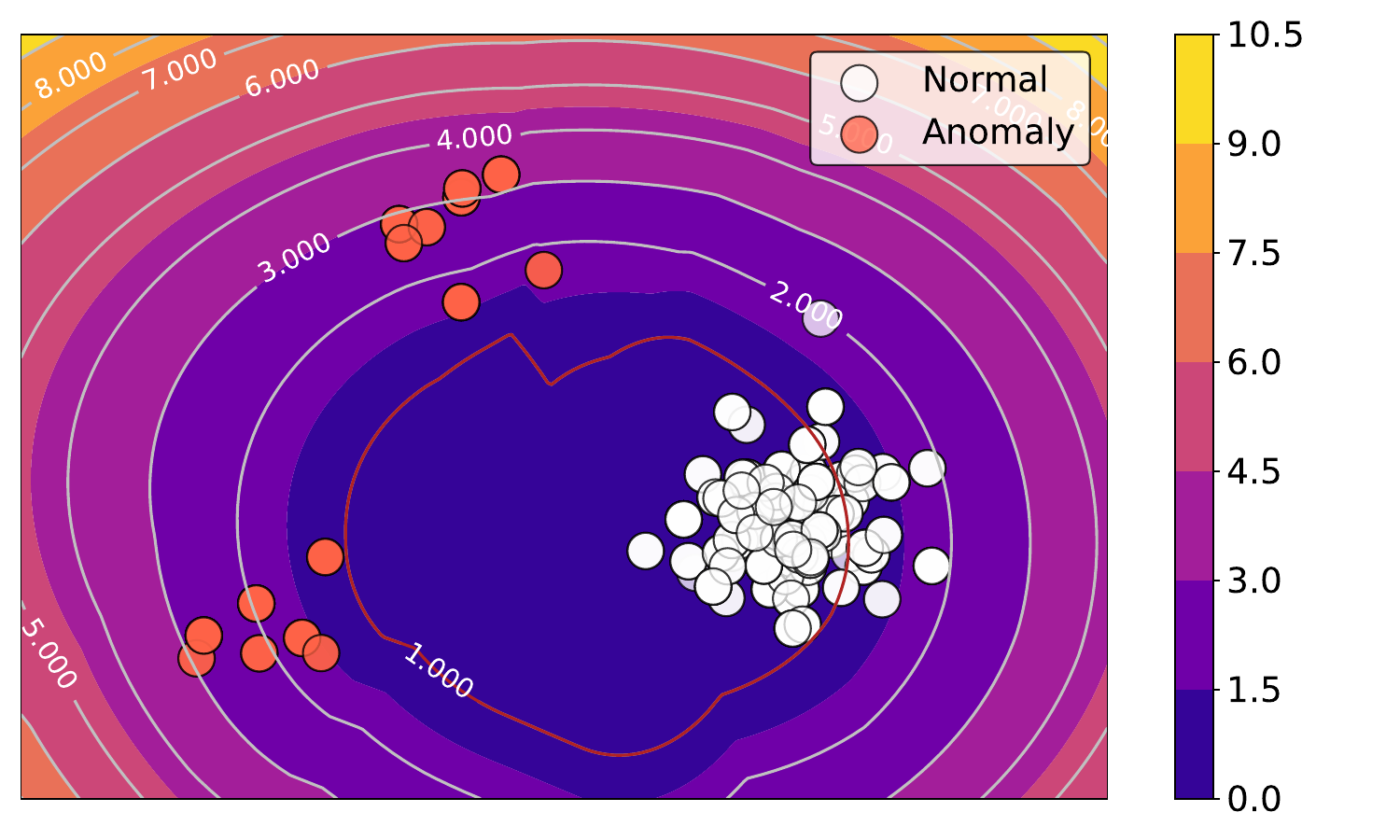}
        \caption{}
        \label{fig:toy_dae_c_zero}
    \end{subfigure}
    \hfill
    \begin{subfigure}[]{.3\textwidth}
    \includegraphics[height=3.5cm]{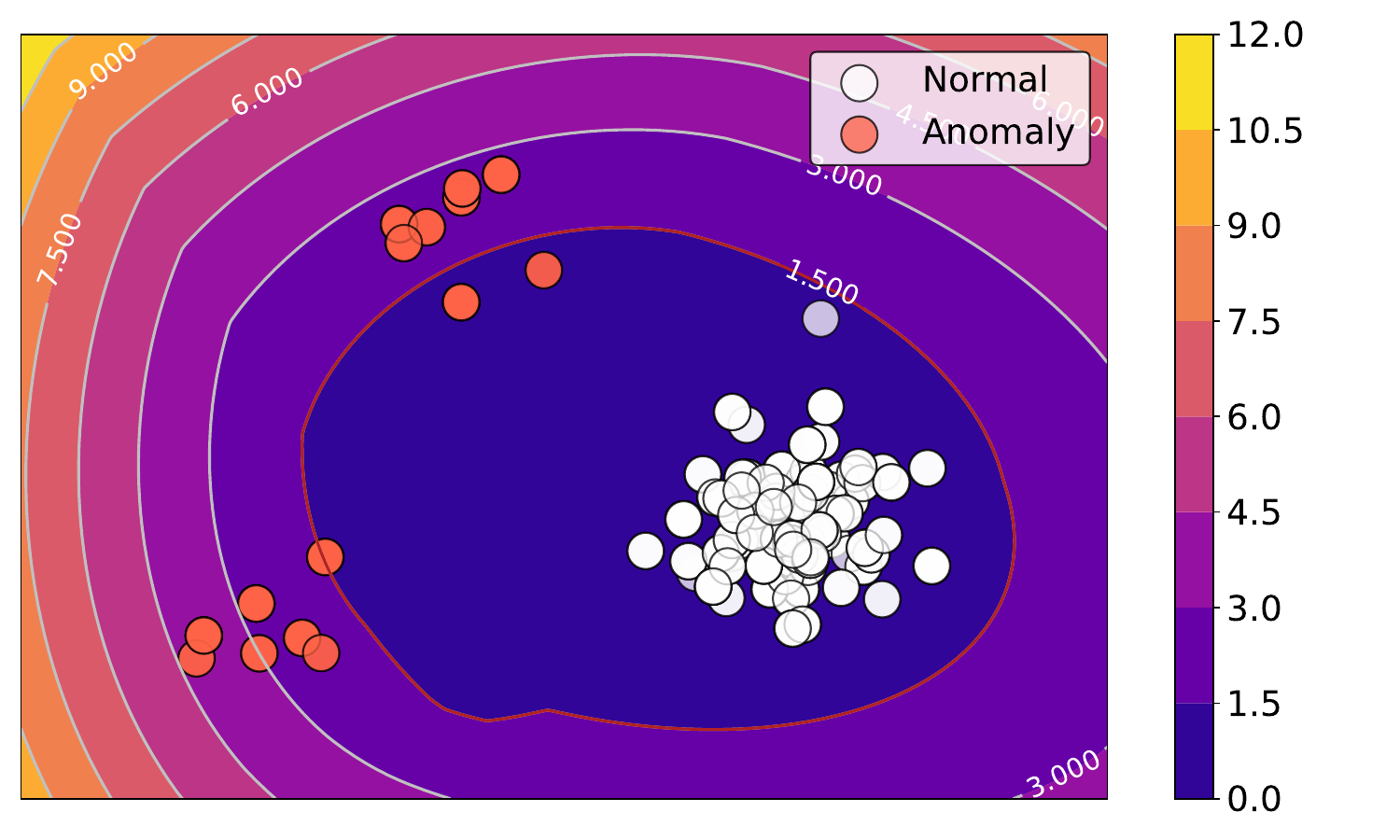}
    \caption{}
    \label{fig:toy_dae_c_learnable}
    \end{subfigure}
    \caption{Contour plot showing the decision boundary of DAE-LR utilizing our proposed loss function from Equation (\ref{eq:recerr_prime}), and trained on 2D synthetic contaminated data with different types of center \( c \). (a) Standard DAE ($\lambda=0$). (b) DAE-LR mean-centered ($c=\Bar{z}$). (c) DAE-LR with a learnable center.}
\end{figure*}

The standard version of DAE allows flexibility in how the model represents data in the latent space, focusing primarily on minimizing the reconstruction error. However, when dealing with contaminated samples, controlling this latent space projection is crucial. Ideally, normal data should be clustered densely in the latent space, making the representation of anomalies more dispersed. To achieve this, we introduce an additional constraint (\(\|z - c\|^2 < \epsilon\)) to regularize the latent representation.

% The standard version of DAE leaves freedom to the model on how it represents data in the latent space; the focus is put only on the reconstruction error. However, in the presence of contaminated samples, we need to control this projection; we would need normal data to be grouped in a dense region in latent space, making anomalies representation scattered. Therefore, to enhance the performance of the DAE, we introduce an additional constraint ($\|z - c\|^2 < \epsilon$) to regularize the latent representation constraint.  

% The standard version of DAE allows flexibility in how the model represents data in the latent space, focusing primarily on minimizing the reconstruction error. However, when dealing with contaminated samples, it is crucial to control this latent space projection. Ideally, normal data should be clustered densely in the latent space, making the representation of anomalies more dispersed. To achieve this, we introduce an additional constraint (\(\|z - c\|^2 < \epsilon\)) to regularize the latent representation.

The optimization formulation for the DAE-LR is then given by:
\begin{equation}
\label{eq:daelr_formulation}
    \min_{\theta} \|x - \hat{x}\|^2 \quad \text{subject to} \quad \|z - c\|^2 \leq \epsilon,
\end{equation}
where \( x \) is the input data, \( \hat{x} \) is the reconstructed data, \( z \) is the latent representation, \( c \) is a learnable center in the latent space, and \( \epsilon \) is a small threshold that controls the clustering of normal data around \( c \).

The problem formulated in Equation \eqref{eq:daelr_formulation} leads to the following loss function, which should be minimized with respect to the parameters \(\theta\) and \(c\):
 \begin{equation}
   \label{eq:recerr_prime}
     L^{\text{(ours)}}_\theta(x) = \|x - \hat{x}\|^2 + \lambda\|z - c\|^2 \,.
 \end{equation}
 where, $z =$ encoder$(x)$ is the latent representation of the input data, $\theta$ and $c$ are models' parameters, and $\lambda > 0$ the regularizing factor for the latent representation. 

%This regularization is incorporated alongside the standard reconstruction error, as outlined in Equation (\ref{eq:recerr_prime}). 

Notably, the combination of latent constraint term and reconstruction error helps control how the input data is represented in the latent space, enforcing it to be centered around a center $c$. Figure \ref{fig:toy_dae_c_learnable} shows how this combination contributes to a better decision boundary compared to the standard version in Figure \ref{fig:toy_dae_standard}, when the training data is contaminated.

 \begin{figure}[ht]
    \centering
    \includegraphics[scale=.23]{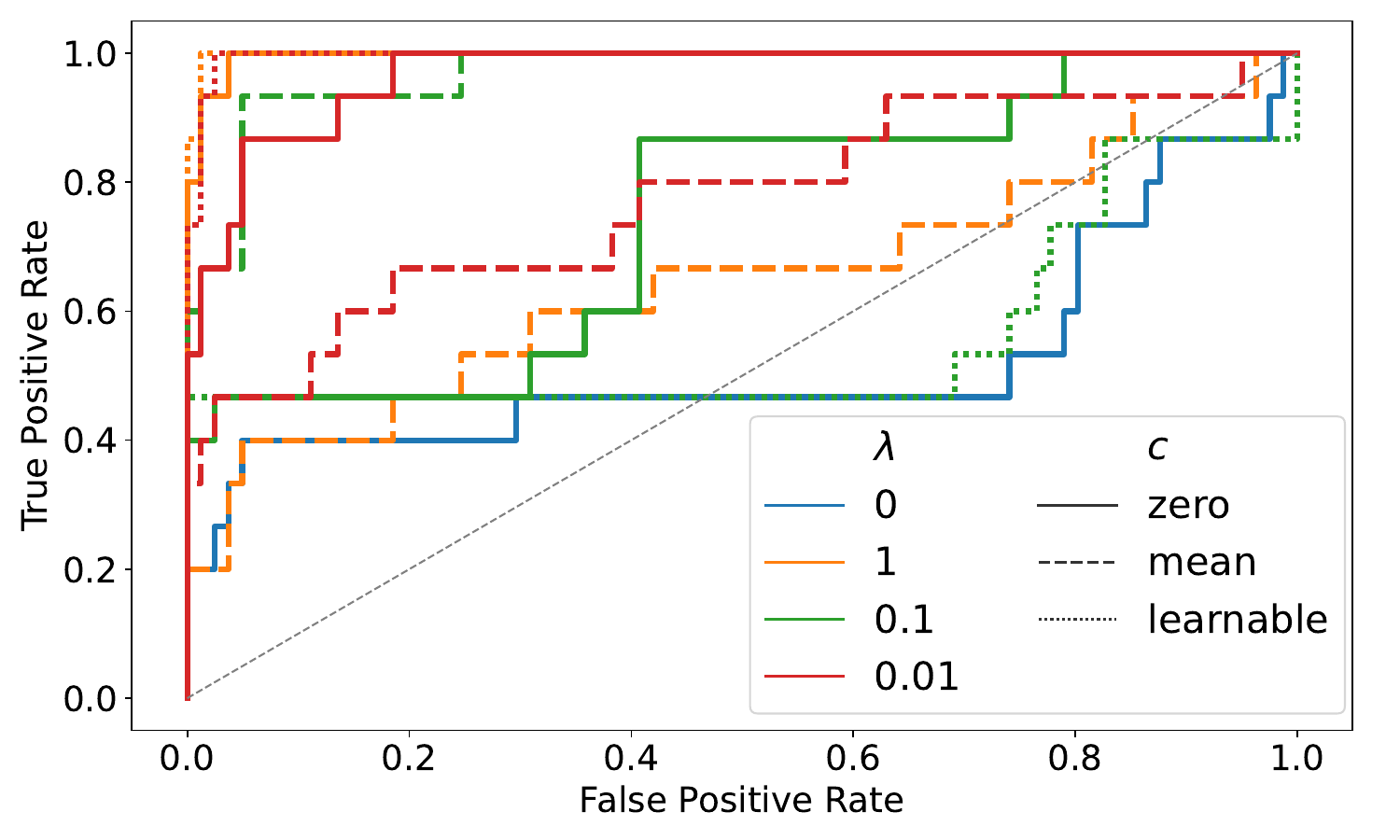}
    \caption{ROC curves illustrating the performance of the latent-regulated DAE-LR (as described in Equation \ref{eq:recerr_prime}), which has been trained on 2D synthetic and contaminated data.  It's important to note that when $\lambda=0$, it corresponds to the standard DAE version.}
    \label{fig:toy_roc}
\end{figure}

% \begin{figure*}[ht]{}
%          \centering
%          \subfloat[]{\label{fig:toy_roc}\includegraphics[height=2.5cm]{./assets/newplot/roc_toy_data}}
%  % \hspace{1pt}
%          \subfloat[]{\label{fig:toy_dae_cplot_zero}\includegraphics[height=2.5cm]{./assets/newplot/dae_rn_0_zero_0}}
%           % \hspace{1pt}
%          \subfloat[]{\label{fig:toy_dae_cplot_center}\includegraphics[height=2.5cm]{./assets/newplot/dae_rn_1_learnable_1}}
%         \caption{Deep Autoencoder with a regulated latent representation (Equation \ref{eq:recerr_prime}) trained on 2D synthetic contaminated data with varying values of $\lambda$ and the type of center $c$ (Note $\lambda=0$, corresponding to the standard DAE version). (a) ROC curves. (b) Contour plot showing the decision boundary for the standard DAE  (i.e., $\lambda=0$). (c) Contour plot of DAE trained using our proposed loss function in Equation (\ref{eq:recerr_prime}).}
% \end{figure*}

 {There are various choices for the center, \( c \). A trivial choice could be to set \( c = \textbf{0} \), or to use the average of latent representations (i.e., \( c = \Bar{z}=\frac{1}{n} \sum^{n}_{i=1} z_i \)). Instead of making such strong assumptions on what the center $c$ should be, we allow the model to determine the optimal center during training. The center \( c \) is a learnable parameter.}

Figure \ref{fig:toy_dae_c_zero} and \ref{fig:toy_dae_c_learnable} display the decision boundary using a mean-centered and learnable-centered latent representation, respectively. We observe that using mean-centered may lead to increasing false positive alerts, while the learnable-centered $c$ produces a better decision boundary, confirming our intuition. Furthermore, Figure \ref{fig:toy_roc} displays ROC curves with different choices of $\lambda$ and the center $c$. We can see that a learnable $c$ yields better performance.

\section{Proposed Evaluation Protocol}
\label{evprot}
\begin{figure*}[!ht]
    \centering
    \includegraphics[height=7cm]{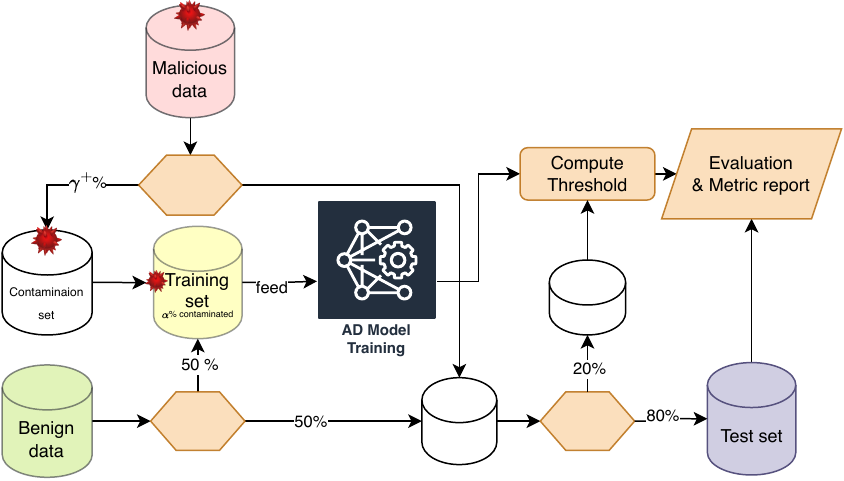}
    \caption{Proposed evaluation protocol workflow for a single run.
    The model is trained with a contaminated set characterized by contamination ratio $\alpha$. The proportion of benign traffic and attack data in the test set is maintained constant across runs, regardless of the level of training set contamination. Moreover, the model threshold for the anomaly scoring function is computed on a separate validation set, distinct from the final test set. The final test set is excluded from the training process to prevent data leakage.}
    \label{fig:datasplitstrategy}
\end{figure*}

This section presents the protocol outlined in Algorithm \ref{alg:anomaly_detection}, which is used to evaluate the robustness of the selected methods. We aim to address the following question concerning our proposed evaluation protocol: \textit{How can we effectively ensure a fair and rigorous model evaluation under contamination, capture variance across runs, mitigate the risk of data leakage, and accommodate the class imbalances typically observed in network intrusion detection datasets?} Furthermore, we investigate the added value of incorporating critical difference diagrams to enhance model performance ranking and facilitate comprehensive pairwise comparisons in this specific context. Figure \ref{fig:datasplitstrategy} illustrates the workflow for one run using this evaluation protocol.

%run, showcasing the data split strategy and threshold computation for assessing the robustness of network AD models against data contamination

% \begin{itemize}
%     \item \djefftodo{}
%      \item \djefftodo{Captured Variance. (multiple seed)}
%      \item \djefftodo{Attack as the positive class (class of interest}
%     \item \djefftodo{Avoid data leakage.}
%     \item \djefftodo{Statistical methods for ranking models..}
% \end{itemize}

\textbf{Data split strategy.} We split the dataset following the setup proposed by \cite{zong2018deep}, the training set contains a proportion $\gamma_{-}$ of normal data samples randomly drawn, but we add a ratio $\alpha$ of \textit{attack related data}. The test set contains the rest of the normal data plus a percentage $(1-\gamma_{+})$ of attack data with $\gamma_{+} \in \{20, 40\}$. The remaining $\gamma_{+}$ percent of attack data is in a set that we call the \textit{contamination} set. More specifically, the ratio $\alpha$ of attack data utilized to contaminate the training set stems from that contamination set.  

The underlying motivation for employing this specific data split strategy is to maintain both consistency between various sets of experiments and fairness in the evaluation of diverse algorithms, as highlighted in \cite{alvarez2022revealing}. By maintaining a constant proportion of normal and attack data within the test set, regardless of the training set contamination level, we facilitate a robust assessment of how varying levels of contamination in the training set impact the performance of models. 

\textbf{Captured Variance.} It's important to note that we consistently apply this data split strategy in every run, with different seeds for randomness, allowing us to capture the variance arising from data sampling, in addition to the variance attributed to model's parameter initialization, as discussed in \cite{bouthillier2021accounting}. 

% \textbf{2. Class of interest.} Since cybersecurity datasets have a significant class imbalance, we consider attack data as the class of interest or the positive class. Attack data is often the minority class, while normal data is the majority class in a real-world application. The scarcity of attack data makes it crucial to evaluate the performance of models on this data. Furthermore, assessing model performance with normal data as the class of interest can overestimate the classifier and be misleading \cite{saito2015precision, alvarez2022revealing}.

\textbf{Performance metrics and threshold.} We report the average and standard deviation of \textit{F1-score, precision, and recall} over 20 runs for every ratio $\alpha$ of the training set contamination. These metrics are suitable for problems with imbalanced class datasets \cite{chicco2020advantages}. 
Since network intrusion detection datasets have a significant class imbalance, we consider attack related data as the class of interest or the positive class. Attack data is often the minority class, while normal data is the majority class in a real-world application. The scarcity of attack data makes it crucial to evaluate the performance of models on this data. Furthermore, assessing model performance with normal data as the class of interest can overestimate the classifier and be misleading \cite{saito2015precision,alvarez2022revealing}.
Therefore, all metrics are computed using the attack data class as the positive class. 
However, F1-score, precision, and recall calculation require setting a decision threshold on the samples' anomaly scores. For that we propose to use 20\% of the test set to find the decision threshold and then test it on the remaining 80\%. We choose the decision threshold such that the F1-score -- the harmonic mean of precision and recall -- is the best that the model could achieve. 

\textbf{Rigourous and leakage free model evaluation.}
It is intuitive that creating disjoint subsets of data as described in the data split strategy mentioned above prevents some sort of data leakage during the determination of the model's threshold or the final evaluation. Subsets created are training, contamination, and test sets. This strategy is data-leakage-free because there is no interaction with the final test set (which represents 80\% of the test set) during the training phase. Such segmentation of data makes the model to be built on the training set with contamination data stemming from the contamination set, and the threshold defined using the 20\% of the test set; the final test set remains entirely untouched and unknown to the model before the final evaluation. This ensures that the model's performance is measured accurately on previously unseen data. It also eliminates any potential bias that may arise if the test set were used to adjust the model's parameters or to determine the model's threshold. Hence, the proposed evaluation protocol indicates that the model's performance and robustness are evaluated objectively.

\textbf{Models comparison and ranking.} We additionally suggest the use of the critical difference diagram (CD diagram) \cite{demvsar2006statistical} to derive the overall model performance ranking and to perform pairwise comparisons of models from a statistical standpoint. The CD diagram is constructed using the Wilcoxon-Holm method. The Wilcoxon-Holm method, which is a pairwise rank test, is recommended when distributions of data being compared are not necessarily normal, and models are run from the same initial conditions which makes pairwise tests more appropriate. In addition, as mentioned above, the Wilcoxon-Holm method uses $p$-values to measure significant differences in data. However, it is mentioned that the $p$-value alone is not enough to describe the statistical difference between two groups of data \cite{effectsize} as one might not be able to quantify, from $p$-values only, how large statistical differences are. Another important metric to measure the significance of differences from two groups of samples needs to be considered\textemdash effect size. In other words, the $p$-value and the effect size should be combined for a more complete statistical conclusion.

\begin{algorithm}[!ht]
% \tiny
\fontsize{8}{10}\selectfont
        \caption{Evaluation Protocol}\label{alg:anomaly_detection}
        \begin{algorithmic}
        \Require $X$, $\gamma_{-}$, $\gamma_{+}$, $g_\Theta$, $\alpha_0$, and $\alpha_1$ represent the dataset, the proportion of normal data for the training set, the proportion of anomalous data for the contamination set, an increment of contamination rate, and the maximum value of the contamination rate, respectively.
\Function{eval-protoc}{$X, \gamma_{+},\gamma_{-}, g_\Theta, \alpha_0,\alpha_1$}
                    \State $X^+\gets $ attack-traffic($X$)
                    \State $X^-\gets $ benign-traffic($X$)  
                    \State metrics $\gets \{$recall, precision, f1-score$\}$
                    \State $N^+\gets \#X^+$
                    \State $N^-\gets \#X^-$
                    \State $M^+\gets \lceil(1-\gamma_{+}) N^+ \rceil$
                    \State $M^-\gets \lceil (1-\gamma_{-}) N^- \rceil$
                    \State $X^+_\text{train}\gets$ random-selection($X^+, M^+$)
                    \State $X^-_\text{train}\gets$ random-selection($X^-, M^-$)
                    \State $X^+_\text{test}\gets X^+\setminus X^+_\text{train}$ 
                    \State $X^-_\text{test}\gets X^-\setminus X^-_\text{train}$ 
                    \State $X_\text{test}\gets X^+_\text{test} \cup X^-_\text{test}$ 
                    \State $X_\text{threshd}\gets$ random-selection($X_\text{test}, \lceil.2*\#X_\text{test}\rceil$)
                    \State $X_\text{test}\gets X_\text{test}\setminus X_\text{threshd}$ 
                    \State $D^+_\text{train}, D^-_\text{train}\gets$preprocessing($X^+_\text{train}, X^-_\text{train}$)
                    \State $D_\text{test}, D_\text{threshd}\gets$preprocessing($X_\text{test}, X_\text{threshd}$)
                    \State $n \gets \#D^-_\text{train}$
                    \State $\alpha \gets 0$
                    % \State significances$\gets \emptyset$
                    % \For{metric $\in$ metrics} 
                    \State scores$\gets [~]$
                    \While{$\alpha  \leq \alpha_1$}
                    \State $m\gets \lceil \frac{\alpha n}{(1-\alpha)}\rceil$
                    \State $C \gets$ random-selection($D^+_\text{train}, m$)
                    \State $D_\text{train} \gets D^-_\text{train} \cup C$
                    \State $g_\Theta\gets $train($g_\Theta, D_\text{train}$)
                    \State threshold $\gets $\text{estimate\_threshold}($g_\Theta, D_\text{threshd}$)
                    \State score $\gets $evaluation($g_\Theta, D_\text{test}$, metric, threshold)
                    \State scores$.push$((score, $\alpha$))
                    \State $\alpha \gets \alpha + \alpha_0$
                    \EndWhile
                    \State \textbf{return} scores
                \EndFunction
    
        \end{algorithmic}
      \end{algorithm}

\section{Experimental investigation}
\label{experiements}
\begin{figure*}[ht]
\centering
\begin{subfigure}[]{.3\textwidth}
    \includegraphics[height=4.5cm]{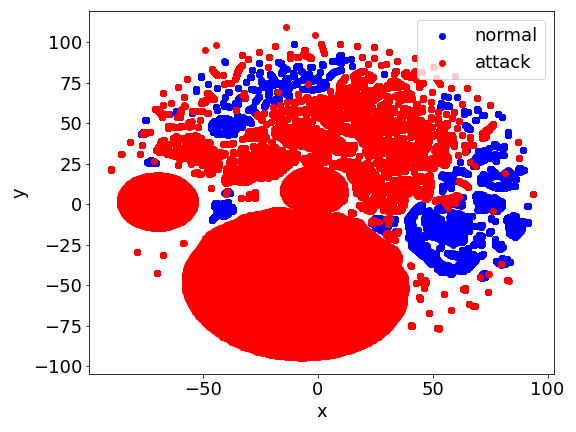}
    \caption{}
    \label{fig:tsnekdd}
\end{subfigure}
\hfill
\begin{subfigure}[]{.3\textwidth}
    \includegraphics[height=4.5cm]{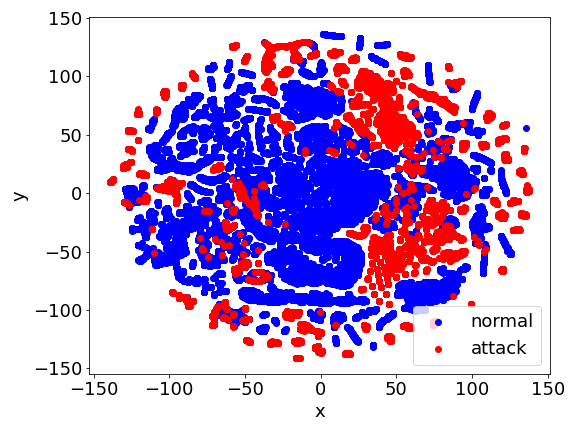}
    \caption{}
    \label{fig:tsnensl}
\end{subfigure}
\hfill
\begin{subfigure}[]{.3\textwidth}
    \includegraphics[height=4.5cm]{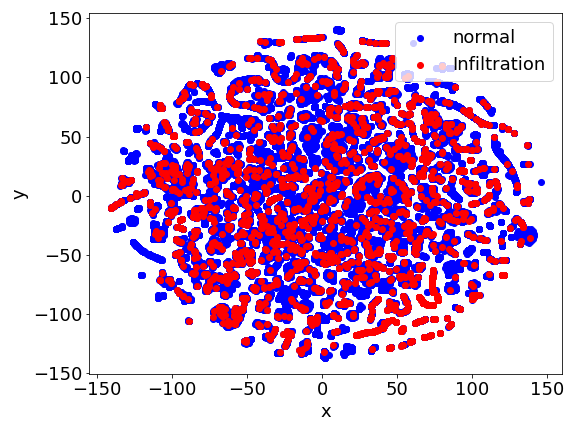}
    \caption{}
    \label{fig:tsnecic}
\end{subfigure}
% \subfloat[]{}
% \hspace{1pt}
% \subfloat[]{\}
% \hspace{1pt}
% \subfloat[]{ }

\caption{Visualization of Normal Traffic and Attack Data in a Two-Dimensional Space Using t-SNE \cite{van2008visualizing}. (a) and (b) Show normal and attack data from the KDDCUP and NSL-KDD datasets, respectively, with all attack types combined into a single class. (c) Displays normal data and specifically highlights the \textit{infiltration} attack data from the CSE-CIC-IDS2018 dataset.}
\label{fig:tnseviz}
\end{figure*}

We first describe the datasets and the models, and then discuss experimental results. The distributions of these datasets are visualized in Figure \ref{fig:tnseviz}, and their characteristics are summarized in Table \ref{dataset-info}.

\subsection{Datasets}
Table \ref{dataset-info} shows the statistics of datasets. We applied Min-Max scaling to the continuous features and one-hot encoding to the categorical features for all the datasets.
% \begin{itemize}

\textbf{KDDCUP}\footnote{\href{http://kdd.ics.uci.edu/databases/kddcup99/kddcup99.html}{http://kdd.ics.uci.edu/databases/kddcup99/kddcup99.html}} is a cyber-intrusion detection dataset that is 20 years old. While it conveys well-known issues such as duplicated samples \cite{tavallaee2009detailed},  it is still widely used as a benchmark dataset. It contains data related to normal traffic and attacks simulated in a military network environment. The Attacks include DoS, R2L, U2R, and probing. The dataset has 41 features, with 34 continuous, and the rest are categorical features. \textbf{NSL-KDD} is a revised version of the KDDCUP dataset provided by the Canadian Institute of Cybersecurity (CIC).
    
    %It counts 80\% of attack data and 20\% of normal data, unlike what one could expect. So, we swapped data labels to reflect a real-world scenario where the majority class is the class of normal data. We use the version that holds only 10 percent of the original data.
    
    %\item \textbf{NSL-KDD} is a revised version of the KDDCUP dataset provided by the Canadian Institute of Cybersecurity (CIC). Some inherent issues, such as duplicated samples from the original KDD dataset, are solved in this version. Further details on this dataset are discussed in \cite{tavallaee2009detailed}.
    
    \textbf{CIC-CSE-IDS2018} is also provided by CIC in collaboration with the Communication Security Establishment. Unlike the KDDCUP dataset, this dataset is recent, and it contains normal traffic plus attacks data simulated on a complex network. Attack types include Brute-force, Heartbleed, Botnet, DoS, DDoS, Web attacks, and infiltration of the network from the inside \cite{sharafaldin2018toward}. In our experiments, we utilized both the original and the revised version of this dataset, referred to as CIC-IDS2018R \cite{9947235} in the experiment section.

     \textbf{Kitsune}\footnote{\href{https://archive.ics.uci.edu/dataset/516/kitsune+network+attack+dataset}{https://archive.ics.uci.edu/dataset/516/kitsune+network+attack+dataset}} is a cybersecurity dataset collected from a commercial IP-based surveillance system. This dataset is related to network traffic within an IoT (Internet of Things) environment. The types of attacks in this dataset include OS scans, fuzzing, video injection, SSDP flooding, and Botnet malware.

    \textbf{CIC-IoT23} is a novel network traffic dataset designed specifically for IoT environments, serving as a benchmark for research and evaluation. The IoT network comprises a total of 105 devices, as detailed in \cite{s23135941}. Here the types of attacks it contains: DDoS and DoS, Recon, Brute Force, Spoofing, and Mirai. %To ensure relevance to real-world contamination scenarios, we preprocessed the dataset differently. Specifically, we performed random downsampling of the DDoS and DoS attack data. This step was taken because these attacks are comparatively easier to detect, and their overrepresentation could potentially skew the evaluation results. 
% \end{itemize}
\begin{table}[h]
\tiny
   \caption{Datasets statistics.}
    \label{dataset-info}
    \centering
    % \tiny
    \begin{tabular}{l|c|c|c}
        \hline
         Dataset &  \# Samples & \# features & Attacks ratio\\
         \hline
         KDDCUP 10\% & 494 021 & 42 & 0.1969\\
         NSL-KDD & 148 517 & 42 & 0.4811\\
         CSE-CIC-IDS2018 & 16 232 944 & 83 & 0.1693\\
          Kitsune & 25 758 272 & 115 & 0.2304\\
          CIC-IoT23 & 4 671 520 & 41 & 0.2704\\
         \hline
    \end{tabular}
    
\end{table}

%Note that, we combined all types of attacks data samples into one class, i.e., \emph{``attack''} for all datasets. Especially for the CSE-CIC-IDS2018 dataset, we kept the original labels to investigate how the contamination of training data with some filtered types of attacks affects each class. Table \ref{dataset-info} shows the statistics of the three datasets. We applied Min-Max scaling to continuous features and one-hot encoding to categorical features for the three datasets.

%Note that we combined all types of attack data samples into a single class, i.e., \emph{``attack''} for all datasets. In particular, for the CSE-CIC-IDS2018 dataset, we kept the original labels to investigate how contamination of the training data by certain filtered attack types affects the detection rate of each class. 

\subsection{Baseline Models}
% \section{Models Description}

We implemented and investigated six unsupervised AD models. Our source code can be found at \url{https://anonymous.4open.science/r/network_anomaly_detection_robustness-4EF3/README.md}.

% http://kdd.ics.uci.edu/databases/kddcup99/kddcup99.html
 
\textbf{Deep Unsupervised Anomaly Detection \cite{li2021deep}}. DUAD is a method that employs a Deep Autoencoder (DAE) for AD. It is based on the hypothesis that anomalies in the training set are samples with relatively high variance distribution. Unlike a standard DAE, DUAD applies distribution clustering after a fixed number of iterations to select a subset of normal data from the training set—specifically, clusters with low variance. The reconstruction error is used as the anomaly score. We integrated the DUAD approach with our DAE-LR model, which we refer to as DUAD-LR.

\textbf{Deep Structured Energy Based Models for Anomaly Detection \cite{zhai2016deep}}. DSEBM, as its name suggests, is an energy-based model. It learns the energy function of input data through a neural network. The algorithm provides two scoring functions based on energy and another based on the reconstruction error. For our experiments, we consider the energy-based anomaly scoring function, where samples with high energy are classified as anomalies.

\textbf{Deep Auto Encoding Gaussian Mixture Model \cite{zong2018deep}}. DAGMM consists of two neural networks: an autoencoder and an estimation network. These networks are trained in an end-to-end fashion. The model concatenates the output of the encoder -- the latent representation of the input -- and the reconstruction error, then feeds them to the estimation network, whose output is used to compute the parameters of a Gaussian Mixture Model (GMM). Specifically, the estimation network is used to obtain sample likelihood, which is also considered the anomaly score.

\textbf{Adversarially Learned Anomaly Detection \cite{zenati2018adversarially}}. ALAD extends Bi-GAN \cite{gupta2022analysis} architecture by adding two more discriminators to ensure data-space and latent-space cycle consistencies. The scoring function is the reconstruction error on the output of an intermediate layer of one of the discriminators.

\textbf{Neural Transformation Learning for Deep Anomaly Detection Beyond Images \cite{qiu2021neural}}. NeuTraLAD is a self-supervised learning method for AD. It combines contrastive learning with the idea of learning data transformation through neural networks to do data augmentation with tabular data. The loss function is defined to maximize agreement between an input and its transformations while minimizing agreement between transformations of an input. The same deterministic loss function is used as the scoring function.

%%%%%%%%%%%%%%%%%%%%%%%%%%%%%%%%%%%%%%
%%    HYPERPARAMETERS               %%
%%%%%%%%%%%%%%%%%%%%%%%%%%%%%%%%%%%%%%
\subsection{Results and Discussion}
Table \ref{f1_results_table} reports F1-score, precision, and recall of the models with different values of contamination ratio $\alpha$.

% Table \ref{f1_results_table} and \ref{f1_results_table_iot} report F1-score, precision, and recall of the models with different values of contamination ratio $\alpha$.

\begin{figure*}[ht]
\centering
\includegraphics[height=4.1cm]{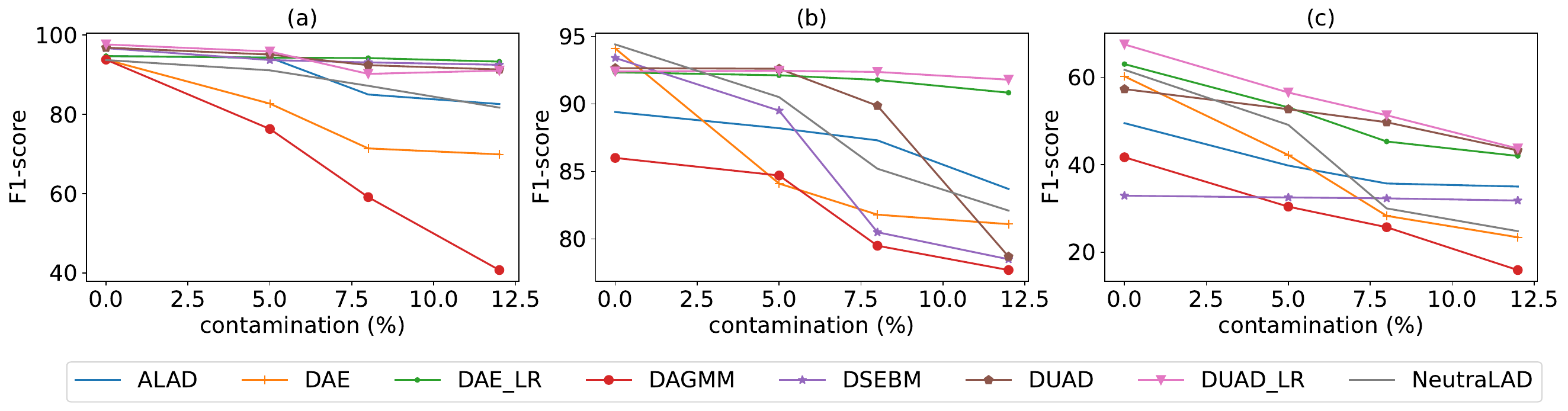}
\caption{Visualization models' average F1-scores, with the training data contamination ratio on the x-axis. (a) For KDDCUP dataset. (b) For NSL-KDD dataset. (c) For CSE-CIC-IDS2018 dataset.}
\label{fig:f1s}
\label{fig:f1cic}
\label{fig:f1nsl}
\label{fig:f1kdd}
\end{figure*}
\begin{table*}[h!]
\caption{Average Precision, Recall, and F1-Score (all with standard deviation) of the models trained exclusively on normal samples contaminated with a ratio $\alpha$ of attacks samples.}
\label{f1_results_table}
% \vskip 0.15in
\centering
\tiny
\begin{tabular}
{@{}l|c|lllllllllllll@{}}
\toprule 
\multirow{2}{*}{Model} & \multirow{2}{*}{$\alpha$} & \multicolumn{3}{c}{KDDCUP 10} & \phantom{a} & \multicolumn{3}{c}{NSL-KDD} & \phantom{a} & \multicolumn{3}{c}{CSE-CIC-IDS2018} \\
\cmidrule{3-5} \cmidrule{7-9} \cmidrule{11-13} 
&  & Pr  & R & $F_1$ && Pr & R & $F_1$ && Pr & R & $F_1$  \\
\hline

                {ALAD}   & \multirow{7}{*}{0\%} & 92.5$\pm$0.9 & 97.2$\pm$5.5 & 94.7$\pm$3.2 && 87.1$\pm$2.5 & 91.9$\pm$1.1 & 89.4$\pm$1.4          && 47.6$\pm$8.4 & 52.0$\pm$5.8 & 49.5$\pm$6.6\\
              {DAGMM} &  & 91.2$\pm$4.9 & 96.7$\pm$3.0 & 93.8$\pm$3.6                        && 86.4$\pm$4.2 & 85.9$\pm$2.4 & 86.0$\pm$1.8          && 42.4$\pm$11.6 & 41.3$\pm$10.1 & 41.7$\pm$10.8\\
              {DSEBM} &  & 94.1$\pm$0.3 & 99.4$\pm$0.1 &{96.7$\pm$0.1}               && 93.6$\pm$0.3 & 93.1$\pm$0.4 & 93.4$\pm$0.1                  && 29.6$\pm$0.1 & 37.0$\pm$0.4 & 32.9$\pm$0.2\\
           {NeuTraAD} &  & 88.2$\pm$0.4 & 99.9$\pm$0.1 & 93.7$\pm$0.2                        && 95.0$\pm$0.9 & 93.8$\pm$0.7 & \textbf{94.4$\pm$0.3} && 69.2$\pm$4.0 & 55.7$\pm$0.9 & 61.7$\pm$1.4\\
            {DAE}  &  & 88.8$\pm$0.4  & 99.4$\pm$0.4  & 93.8$\pm$0.1                      && 95.3$\pm$1.7 & 93.2$\pm$1.2 & 94.2$\pm$0.3          && 61.9$\pm$8.2 & 59.5$\pm$5.6 & 60.2$\pm$4.3\\
               {DUAD} &  & 94.3$\pm$0.2 & 99.5$\pm$0.0 & 96.8$\pm$0.1                        && 92.2$\pm$0.2 & 93.2$\pm$0.3 & {92.6$\pm$0.1}         && 56.5$\pm$0.3 & 58.2$\pm$0.1 & 57.3$\pm$0.1\\
            \textbf{DAE-LR}  &  & 90.0$\pm$0.1 & 99.9$\pm$0.1 & 94.7$\pm$0.1                        && 92.3$\pm$0.1 & 92.5$\pm$0.1 & 92.4$\pm$0.0          && 69.5$\pm$3.6 & 57.6$\pm$0.8 & 63.0$\pm$1.5\\
            \textbf{DUAD-LR} &  & 96.0$\pm$0.2 & 99.3$\pm$0.0 & \textbf{97.6$\pm$0.1 }              && 92.3$\pm$0.1 & 92.5$\pm$0.1 & 92.4$\pm$0.0         && 62.2$\pm$3.9 & 73.8$\pm$0.6 & \textbf{67.5$\pm$2.3}\\

\cmidrule{1-13}
               {ALAD} & \multirow{7}{*}{5\%} & 90.6$\pm$4.8 & 98.5$\pm$1.0 & 94.3$\pm$3.0 &     & 88.2$\pm$2.3 & 88.2$\pm$4.2 & 88.2$\pm$2.8 &      & 36.9$\pm$2.4 & 43.4$\pm$5.4 & 39.8$\pm$3.6\\
               
               {DAGMM}  &  & 76.1$\pm$21.0 & 77.3$\pm$21.6 & 76.3$\pm$20.9                  && 83.0$\pm$6.0 & 86.5$\pm$4.6 & 84.7$\pm$5.1           && 31.6$\pm$14.2 & 29.9$\pm$8.4 & 30.4$\pm$11.0\\
               {DSEBM}&  & 88.1$\pm$0.1 & 100.0$\pm$0.0 & 93.7$\pm$0.0                      && 86.0$\pm$1.6 & 93.2$\pm$1.8 & 89.5$\pm$1.7           && 29.2$\pm$0.1 & 36.6$\pm$0.1 & 32.5$\pm$0.1\\
              {NeuTraAD}  &  & 85.1$\pm$0.5 & 98.0$\pm$1.4 & 91.1$\pm$0.7                   && 87.0$\pm$1.8 & 94.3$\pm$1.9 & 90.5$\pm$1.9           && 47.2$\pm$5.4 & 51.9$\pm$7.6 & 49.1$\pm$5.5\\
              {DAE}    &  & 79.6$\pm$7.5  & 86.4$\pm$10.2 & 82.7$\pm$8.3                   && 82.6$\pm$2.6 & 85.5$\pm$4.2 & 83.9$\pm$1.1           && 39.5$\pm$6.0 & 45.9$\pm$8.6 & 42.2$\pm$6.7\\
                {DUAD}  &  & 90.7$\pm$0.0 & 100.0$\pm$0.0 & 95.1$\pm$0.0                    && 92.4$\pm$0.3 & 92.9$\pm$0.2 & \textbf{92.6$\pm$0.0}  && 49.2$\pm$0.3 & 56.9$\pm$0.2 & 52.7$\pm$0.2\\
                \textbf{DAE-LR}  &   & 89.3$\pm$0.1 & 100.0$\pm$0.0 & {94.3$\pm$0.1}                    && 92.1$\pm$0.1 & 92.8$\pm$0.1 & 92.4$\pm$0.0           && 54.9$\pm$5.2 & 52.0$\pm$3.8 & 53.1$\pm$2.2\\
               \textbf {DUAD-LR}  &  & 92.2$\pm$0.2 & 99.8$\pm$0.0 & \textbf{95.9$\pm$0.1 }        && 92.5$\pm$0.9 & 90.9$\pm$1.1 & {91.7$\pm$0.2}  && 51.9$\pm$5.2 & 62.1$\pm$5.9 & \textbf{56.5$\pm$5.4}\\

\cmidrule{1-13}
               {ALAD} & \multirow{7}{*}{8\%} & 80.5$\pm$9.6 & 90.6$\pm$15.8 & 85.0$\pm$12.4 &     & 87.3$\pm$3.9 & 87.5$\pm$2.2 & 87.3$\pm$2.0 &      & 34.9$\pm$7.1 & 36.8$\pm$6.2 & 35.7$\pm$6.1\\
               
               {DAGMM}  &  & 60.4$\pm$11.7 & 59.0$\pm$16.8 & 59.1$\pm$13.4          && 77.9$\pm$9.5 & 81.4$\pm$7.8 & 79.5$\pm$8.3       && 23.7$\pm$10.1 & 28.3$\pm$12.9 & 25.7$\pm$11.3\\
               {DSEBM}   &  & 87.3$\pm$0.2 & 99.6$\pm$0.2 & 93.1$\pm$0.2            && 77.3$\pm$0.7 & 84.0$\pm$0.8 & 80.5$\pm$0.7       && 29.0$\pm$0.1 & 36.5$\pm$0.2 & 32.3$\pm$0.1\\
                
              {NeuTraAD}  &  & 81.5$\pm$2.7 & 94.0$\pm$2.2 & 87.2$\pm$2.1           && 82.5$\pm$1.7 & 88.1$\pm$2.7 & 85.2$\pm$1.6       && 27.3$\pm$13.2 & 33.3$\pm$16.2 & 30.0$\pm$14.5\\
              {DAE}     &  & 68.5$\pm$3.8  & 75.1$\pm$10.2 & 71.4$\pm$5.9          && 80.2$\pm$2.4 & 84.8$\pm$2.8 & 82.4$\pm$0.9       && 25.8$\pm$8.6 & 31.6$\pm$11.8 & 28.3$\pm$9.9\\
                {DUAD}   &  & 87.6$\pm$0.4 & 97.8$\pm$0.3 & 92.4$\pm$0.3            && 88.0$\pm$0.4 & 91.9$\pm$0.7 & 89.9$\pm$0.6       && 45.1$\pm$0.3 & 55.3$\pm$0.4 & 49.7$\pm$0.3\\
               \textbf{DAE-LR}  &   & 89.1$\pm$0.1 & 100.0$\pm$0.0 & \textbf{94.2$\pm$0.0}     && 91.8$\pm$0.1 & 93.0$\pm$0.1 & \textbf{92.4$\pm$0.0}       && 43.9$\pm$2.9 & 47.2$\pm$4.0 & 45.3$\pm$2.5\\
                \textbf{DUAD-LR}   &  & 83.7$\pm$0.3 & 97.8$\pm$0.1 & 90.2$\pm$0.2         && 92.9$\pm$1.3 & 90.3$\pm$1.6 & {91.6$\pm$0.2} && 46.7$\pm$6.8 & 56.9$\pm$8.0 & \textbf{51.3$\pm$7.2}\\

\cmidrule{1-13}
               {ALAD}  & \multirow{7}{*}{12\%} & 77.6$\pm$9.6 & 89.0$\pm$17.3 & 82.6$\pm$13.2 &    & 80.6$\pm$4.3 & 87.0$\pm$4.5 & 83.7$\pm$4.4 &      & 33.0$\pm$4.0 & 37.5$\pm$4.4 & 35.0$\pm$3.7\\
              
               {DAGMM}  &  & 42.8$\pm$13.5 & 39.3$\pm$14.9 & 40.7$\pm$13.9          && 75.9$\pm$14.0 & 79.7$\pm$13.5 & 77.7$\pm$13.6    && 13.4$\pm$7.8 & 20.6$\pm$12.7 & 15.9$\pm$9.0\\
               {DSEBM}    &  & 86.7$\pm$0.4 & 99.1$\pm$0.3 & 92.5$\pm$0.4           && 77.5$\pm$0.4 & 79.5$\pm$0.2 & 78.5$\pm$0.2       && 28.6$\pm$0.1 & 35.7$\pm$0.2 & 31.8$\pm$0.2\\
              {NeuTraAD}  &  & 75.4$\pm$3.2 & 89.4$\pm$2.8 & 81.7$\pm$2.5           && 81.2$\pm$2.7 & 83.2$\pm$3.0 & 82.1$\pm$0.9       && 24.2$\pm$11.8 & 25.8$\pm$14.2 & 24.8$\pm$12.8\\
              {DAE}     &  & 69.0$\pm$7.2 & 71.4$\pm$8.5 & 69.9$\pm$6.2            && 79.7$\pm$1.8 & 83.1$\pm$1.8 & 81.3$\pm$0.5       && 21.0$\pm$5.9 & 26.6$\pm$7.6 & 23.4$\pm$6.5\\
                {DUAD}   &  & 85.9$\pm$0.2 & 97.5$\pm$0.3 & 91.3$\pm$0.2            && 77.1$\pm$0.2 & 80.4$\pm$0.1 & 78.7$\pm$0.1       && 41.3$\pm$0.6 & 45.8$\pm$1.1 & 43.3$\pm$0.8\\
                \textbf{DAE-LR}  &  & 88.3$\pm$0.4 & 98.9$\pm$0.2 & \textbf{93.3$\pm$0.2 }      && 90.8$\pm$0.3 & 92.9$\pm$0.2 & \textbf{91.8$\pm$0.1}       && 43.2$\pm$3.8 & 41.6$\pm$5.3 & 42.0$\pm$2.2\\
                \textbf{DUAD-LR}   &  & 85.4$\pm$0.4 & 97.6$\pm$0.2 & 91.0$\pm$0.4         && 90.8$\pm$1.0 & 92.2$\pm$0.6 & {91.5$\pm$0.4} && 41.2$\pm$0.9 & 46.9$\pm$0.2 & \textbf{43.7$\pm$0.1}\\
% \cmidrule{1-13}
\bottomrule
\end{tabular}
% \caption{Performance metrics on cybersecurity datasets.}
\end{table*}

\begin{figure*}[h]{}
         \centering
%          \subfloat[]{\label{fig:tsnekdd_}\includegraphics[]{}}
% % \hspace{1pt}
% \subfloat[]{\label{fig:tsnensl_}\includegraphics[]{}}
% % \hspace{1pt}
% \subfloat[]{ \label{fig:tsnecic_}\includegraphics[]{}}
         \includegraphics[height=4.1cm]{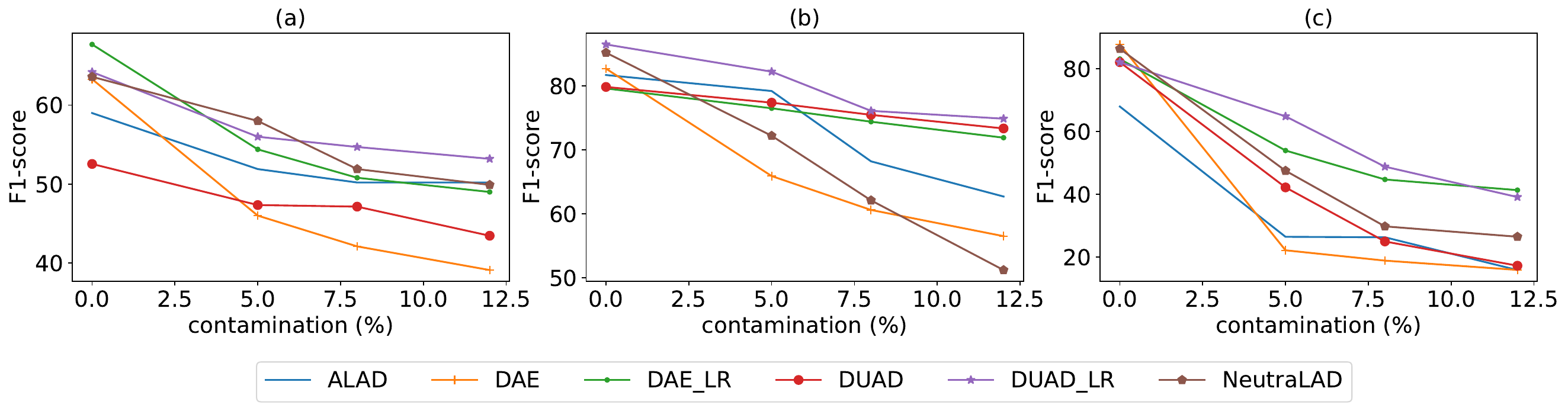}
        \caption{Models' average F1-scores, with the training data contamination ratio on the x-axis. (a) For Kitsune dataset, (b) For CIC-IoT23 dataset, (c) For the revised version of CIC-IDS 2018 dataset. }
        {\phantomsubcaption\label{fig:f1s_iot_a}}
        {\phantomsubcaption\label{fig:f1s_iot_b}}
        {\phantomsubcaption\label{fig:f1s_iot_c}}
        \label{fig:f1s_iot}
\end{figure*}

\textbf{Performance on KDDCUP dataset.} When there is no contamination, all models perform well, with an average F1-score above 90\% across 20 runs. DSEBM and DUAD-LR outperform the other models with an F1-score greater than 96\%. From the two-dimensional representation of a subset of the KDDCUP dataset in Figure \ref{fig:tsnekdd}, it is clear that normal and attack data are mostly distinguishable, with minimal overlap. This may explain the good performance of all models on the KDDCUP dataset. Figure \ref{fig:f1s}a shows performance degradation for all models as the contamination ratio increases, albeit to varying extents. Notably, the F1-scores of DAGMM and DAE decrease dramatically from 93.8\% to 40.7\% and from 93.8\% to 69.9\%, respectively, with a contamination ratio of 12\%. Remarkably, DAE-LR and DUAD-LR maintain good performance even under varying levels of contamination.

DAGMM estimates a Gaussian mixture without explicitly handling outliers during training, causing contaminated samples to influence density estimation and shift normal samples toward low-density regions. However, DUAD and DSEBM show resistance to contamination. DUAD's approach of re-evaluating the training data seems to eliminate some contaminated samples, allowing for training on a nearly clean subset. Even with 12\% contamination, DUAD-LR and DAE-LR demonstrate significant improvements over their standard counterparts.

As for DSEBM, recall remains around 99\%, while precision decreases. This suggests that DSEBM effectively classifies all attacks, assigning high energy to them, regardless of the contamination level. However, DSEBM's precision decreases as contamination in the training data increases."

\textbf{Performance on NSL-KDD dataset.} Figure \ref{fig:f1s}b presents different patterns than those observed with the KDDCUP dataset. Note that NSL-KDD is a relatively balanced dataset. For instance, DSEBM shows poor performance, with its F1-score dropping from 93\% to 78\% at a contamination rate of 12\%. This may be attributed to the removal of duplicates in the KDDCUP dataset. In contrast, DUAD-LR's distribution clustering and latent regularizer play a central role in defending against contamination of the training set. DUAD consistently performs well, maintaining an F1-score of 91\%, regardless of the contamination level. Furthermore, DAE-LR outperformed the other models when faced with 12\% contamination.

\begin{figure}[]
  \centering
     \includegraphics[height=2cm]{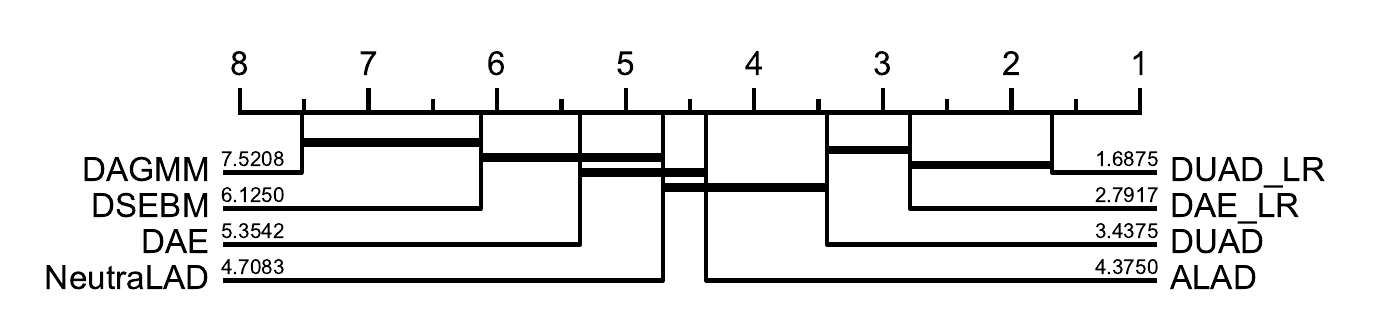}
     \caption{CD-Diagram for overall average rank based on F1-scores of the models on all datasets with different levels of contamination. The lower the ranking, the better the model performs compared to others.}
    \label{fig:ranking}
\end{figure}

\textbf{Performance on Kitsune(IoT) and CIC-IoT23.} 
% {\color{blue} test}
Based on the ranking in Figure \ref{fig:ranking}, we have selected the top models, namely ALAD, DAE, DAE-LR, DUAD, DUAD-LR, and NeutraLAD which we trained on Kitsune and CIC-IoT23 datasets. Figure \ref{fig:f1s_iot} displays their performance under varying levels of data contamination. We observe that NeutraLAD outperforms the others with clean training data by the f1-score of 85.2\% on CIC-IoT23. However, its performance suffered a significant drop of over 30\% when trained with a 12\% contamination rate. This emphasizes the challenge of this self-supervised approach in handling data contamination. On the other hand, DAE-LR and DUAD-LR still better perform with different levels of contamination in both Kitsune and CIC-IoT23. We also note that none of the models was able to achieve an F1-score exceeding 70\% on the Kitsune dataset. This poor performance reveals how hard it is to detect advanced cyber attacks  on this dataset in an unsupervised fashion.

\textbf{Performance on CSE-CIC-IDS2018 dataset.} On this dataset, DUAD-LR achieves the highest F1-score (67\%), followed by DAE-LR (63\%) and NeuTraLAD  (61\%). Overall, all the models struggle to achieve high performance with regards to what they yielded on KDDCUP and NSL-KDD datasets, even without contamination. In Figure \ref{fig:tsnecic}, we present a t-SNE visualization of a subset of normal data and infiltration attack data from the CSE-CIC-IDS2018 dataset. Interestingly, we notice a significant number of overlaps between benign traffic and attack, rendering their distinction challenging. In Figure \ref{fig:f1cic}c, a decrease in F1-scores for all models is observed as the contamination rate increases. While DUAD-LR has previously demonstrated resilience to contamination, its F1-score decreases from 67\% to 43\% on CSE-CIC-IDS2018 with a contamination rate of 12\%. This decline in performance can be attributed to the fact that, unlike KDDCUP and NSL-KDD (as illustrated in Figures \ref{fig:tsnekdd} and \ref{fig:tsnensl}), some of the attack samples in CSE-CIC-IDS2018 (Figure \ref{fig:tsnecic}) exhibit less dissimilarity from normal traffic data. Consequently, eliminating contamination from the training set, even for DUAD-LR, proves to be challenging. Nonetheless, DUAD-LR consistently maintains a higher F1-score compared to other models across different contamination rates, followed by DAE-LR and then ALAD. A similar performance trend is observed on the revised version of this dataset. However, the models exhibit superior performance in the absence of contamination, followed by a steep decline as the contamination ratio increases, as shown in Figure \ref{fig:f1s_iot_c}.

\textbf{Discussion}

This study covers a wide range of approaches in unsupervised AD, showing that generative models like ALAD and self-supervised models like NeuTralAD can achieve great performance in AD tasks, including network intrusion detection. 
The overall ranking of the models is illustrated in Figure \ref{fig:ranking}. The bold horizontal lines connecting groups of models indicate that they are not statistically different based on $p$-values. DUAD-LR, DAE-LR, ALAD, and NeuTralAD emerge as the top-performing models overall. Particularly in scenarios where the training set is contaminated, DUAD-LR and DAE-LR stand out, demonstrating the benefit of regulating the latent representation for AD tasks involving contaminated datasets.  Moreover, results obtained on KDDCUP and NSL-KDD may be misleading when compared to the realities of current computer networks. Models that excel on these outdated datasets may not perform as well on more relevant datasets such as CSE-CIC-IDS2018, Kitsune, and CIC-IoT23, which better simulate the current state of computer networks and cyber threats. Thus, there is a pressing need for more representative network intrusion detection datasets.

\textbf{Future directions.} Based on our experiments, it becomes apparent that data contamination can have a negative impact on the performance of even the most advanced models. Given the unpredictable nature of data cleanliness in real-world scenarios, it is primordial to incorporate defenses against contamination when developing DL models for cybersecurity. One potential defense strategy could involve inferring data labels during the training process. Once inferred labels are available, the algorithm could dynamically identify and remove potentially anomalous instances from the training dataset, enhancing robustness.

We observed that DUAD-LR exhibits consistent resilience to contamination due to its approach of re-evaluating the training set through clustering. However, one drawback of DUAD is its relatively slow training process, especially on large datasets, as it performs clustering over the entire training set multiple times. To address this, one could explore extending the clustering concept introduced by DUAD to identify and reject contaminated data in an online and more computationally efficient manner.

\section{Conclusion}
\label{conclu}
In this rapidly evolving era of cyberattacks, data-driven anomaly detection methods in cybersecurity must demonstrate resilience against data contamination attacks. This paper has contributed valuable insights into the robustness of recent data-driven anomaly detection models when applied to network intrusion detection datasets. Our experiments demonstrate that the latent regulated auto-encoder that we propose in this study, leads to competitive performance when the training data is contaminated. We also observe that even state-of-the-art deep learning models for AD experience a notable drop in performance when the training data is contaminated by attack instances. This underscores the critical significance of robustness to contamination as a pivotal criterion in selecting an anomaly detection model for cybersecurity applications.
Furthermore, our study underscores the potential pitfalls of relying solely on outdated cybersecurity datasets for model evaluation. The experiments conducted reveal that model performance on such datasets may lead to misguided conclusions when applied to the ever-evolving landscape of computer networks and emerging attack patterns.
In conclusion, our findings provide  insights that will inform ongoing efforts in the development of cyberdefense solutions. This includes the imperative need for designing defense frameworks that address training set contamination challenges for deep anomaly detection methods in the context of cybersecurity. These insights are pivotal in enhancing the resilience and effectiveness of anomaly detection systems in the presence of persistent cyberattacks.

%================================================================================

\bibliography{refs.bib}
\bibliographystyle{acl_natbib.bst}

\end{document}